\documentclass{article}

\usepackage{algorithm,algorithmic}
\usepackage{amsfonts,amsmath,amssymb}
\usepackage{graphicx}
\usepackage[colorlinks=true,pdftex]{hyperref} 
\usepackage[utf8]{inputenc}
\usepackage{listings}
\usepackage{listings}
\usepackage{xcolor}

\usepackage{url}
\usepackage{xcolor}
\usepackage{xfrac}

\newcommand{\unit}{\ensuremath{[0,1]}}
\newcommand{\K}{\mathcal{K}}
\renewcommand{\S}{\mathcal{S}}
\renewcommand{\O}{\mathcal{O}}

\newcommand{\nd}{\noindent }
\newcommand{\tuple}[1]{\langle #1 \rangle}
\newcommand{\qed}{\hfill$\square$}
\newcommand{\Name}{\texttt{Q2S2}}

\newtheorem{example}{Example}[section]

\renewcommand{\algorithmicrequire}{\textbf{Input:}}
\renewcommand{\algorithmicensure}{\textbf{Output:}}

\begin{document}

\title{Fuzzy Quantification over OWL Ontologies and Knowledge Graphs}

\author{Enrique Palac\'in$^1$ \and Fernando Bobillo$^1$$^2$ \and Ignacio Huitzil$^1$$^2$ \and Francesca A. Lisi$^3$ \and Umberto Straccia$^4$}

\date{%
    $^1$University of Zaragoza, Zaragoza, Spain\\
    $^2$Aragon Institute of Engineering Research (I3A), Zaragoza, Spain\\
    $^3$University of Bari Aldo Moro, Bari, Italy\\
    $^4$CNR-ISTI, Pisa, Italy
}

\maketitle


\begin{abstract}
This paper presents a versatile framework for evaluating fuzzy quantification queries over both standard and fuzzy ontologies as well as knowledge graphs. The primary objective is the retrieval of individuals that satisfy queries articulated via Type I or Type II fuzzy quantified expressions. A key advantage of the proposed approach is its inherent adaptability: it remains entirely agnostic to the quantifier type, the underlying evaluation method, and the specific data source of the ontology (i.e., OWL ontologies or RDFS knowledge graphs). Furthermore, we present \Name, a publicly accessible implementation of this system developed to support future research.
\end{abstract}


\section{Introduction}


\noindent \emph{Knowledge Representation and Reasoning} (KRR) are foundational pillars of \emph{Artificial Intelligence} (AI), playing a critical role in the development of complex intelligent systems. Currently, the most prominent frameworks for KRR stem from Semantic Web technologies. Notable among these are \emph{OWL ontologies}, which provide formal vocabulary specifications for specific domains~\cite{DLtextbook}, and \emph{Knowledge Graphs}~\cite{kg-book} (KGs), such as \emph{RDFS} graphs~\cite{RDFS}, which offer scalable, graph-based models for capturing extensive datasets. Let us note that the logical counterpart of these two formalisms are \emph{Description Logics} (DLs)~\cite{DLtextbook} and  $\rho$df~\cite{Munoz09,Munoz07}.
Among various extensions of them, let us mention, as relevant to our paper, those that are based on \emph{fuzzy logic}~\cite{Zadeh65}, whose primary aim is to extend these standard, classical logics to represent and reason with vague, imprecise, or uncertain knowledge within structured domains (see, e.g.,~\cite{Bobillo24,MFL,ESWA2012,IJAR2011,KBS2016,Bobillo17,Bobillo18,Straccia09f,Straccia12b,FuzzyOntologies,Straccia14,Zimmermann12}).


For our purposes, it is worth recalling that \emph{quantified sentences} play a critical role in articulating both the knowledge to be represented (e.g., `all humans are mortal') and the specific information to be retrieved (e.g., `retrieve all parents who have a blonde child').
%
%
%
In classical logic and standard DLs, as well as in `almost all' \emph{fuzzy} DL literature~\cite{FuzzyOntologies}, quantification is strictly limited to the existential ($\exists$) and universal ($\forall$) quantifiers. However, this binary paradigm is often overly restrictive, and the ability to express more nuanced quantification is highly desirable in many practical scenarios. Ironically, the assertion that `almost all' papers restrict themselves to these standard quantifiers inherently relies on a fuzzy quantifier itself: specifically, the term `almost all'.

\begin{example}
\label{ex:blackMetal}   
Let us consider an example in the musical domain. Assume that we want to listen to some \emph{black metal}, so we ask an intelligent recommendation system to suggest bands playing this musical style. We can ask it to recommend bands that have released \emph{some} black metal album, by using a query involving the $\exists$ quantifier. However, many bands started playing black metal, but then evolved to other musical styles, which we are not interesting in. 

Another option consists in asking to retrieve those bands such that \emph{all} their albums are black metal, by using a query involving the $\forall$ quantifier. Unfortunately, this query is also not satisfactory. For instance, the Norwegian band \emph{Darkthrone} is widely considered as the black metal band par excellence, but it would not meet the previous restriction, since in the first of their (currently) 23 studio albums they play \emph{death metal} rather than black metal. 

A possible solution may consist of asking to retrieve bands such that \emph{almost all} their albums are black metal and, thus, Darkthrone would satisfy the query with a very high degree. \qed

\end{example}


\nd So, what we propose in this work is a versatile framework for resolving fuzzy quantification queries over semantic knowledge bases, which includes ontologies and knowledge graphs. The primary objective is the evaluation of Type I or Type II fuzzy quantified queries. A key advantage of the proposed approach is its inherent adaptability: it remains entirely agnostic to the quantifier type, the underlying evaluation method, and the specific data source of the ontology (i.e., OWL ontologies or RDFS knowledge graphs). Furthermore, a publicly accessible implementation of this system (called \Name) has been developed to support future research. In particular, on the one hand, we discuss how to solve fuzzy quantification queries over classical ontologies and knowledge graphs, by combining classical knowledge with fuzzy datatypes and fuzzy quantifiers, in such a way that classical semantic reasoners or answering systems can be used. On the other hand, we discuss the case of fuzzy ontologies and fuzzy knowledge graphs, requiring non-standard fuzzy tools.

The remaining of this paper is organized as follows. Section~\ref{sec:background} provides some background. Then, Section~\ref{sec:queries} presents the different queries that will be supported, and Section~\ref{sec:solving} explains how to solve them. Next, Section~\ref{sec:implementation} illustrates our prototype implementation. A detailed comparison of our contribution with other previous work can be found in Section~\ref{sec:relatedWork}. Finally, Section~\ref{sec:conclusions} concludes and points to some ideas for future work.


\section{Background}
\label{sec:background}

\noindent This section provides some minimal background on fuzzy logic (Section~\ref{sec:21}), fuzzy quantification (Section~\ref{sec:22}), and Semantic Web technologies (Section~\ref{sec:23}), needed to follow the rest of the paper.

\subsection{Fuzzy logic}
\label{sec:21}

\noindent \emph{Fuzzy logic} is widely used to manage imprecise and vague knowledge. It is a generalization of classical logic proposed by Zadeh where statements are not necessarily either true or false, but hold to some degree of truth~\cite{Zadeh65}. 

The cornerstone of fuzzy logic is the concept of fuzzy set, which is a generalization of a classical set (also called \emph{crisp}) where elements can have a partial membership. Given a crisp set $X = \{x_1, \dots, x_n \}$ (the \emph{universe}), a \emph{fuzzy set} $A$ is characterized by a membership
function $\mu_{A}(x)$, which associates with each object $x \in X$ a real
number in $[0,1]$ representing the membership degree of $x$ in
$A$. As in classical sets, $0$ means no-membership and $1$ full membership.  But now, an intermediate value between $0$ and $1$ denotes partial membership to $A$. The fuzzy set $A$ may also be represented as $A = \{\mu_A(x_1)/x_1, \ldots , \mu_A(x_n)/x_n \}$. A \emph{normal fuzzy set} is a fuzzy set,  where the maximum membership value (height) is exactly $1.0$. An easy way to normalize a fuzzy set $A$ is by redefining it by means of
\[
\mu_{A_{norm}}(x) = \frac{\mu_A(x)}{\max_{y \in X}\mu_A(y)} \ .
\]

\begin{figure*}[thbp]
\begin{center}
\begin{tabular}{cccc}
\includegraphics[height=2.2cm]{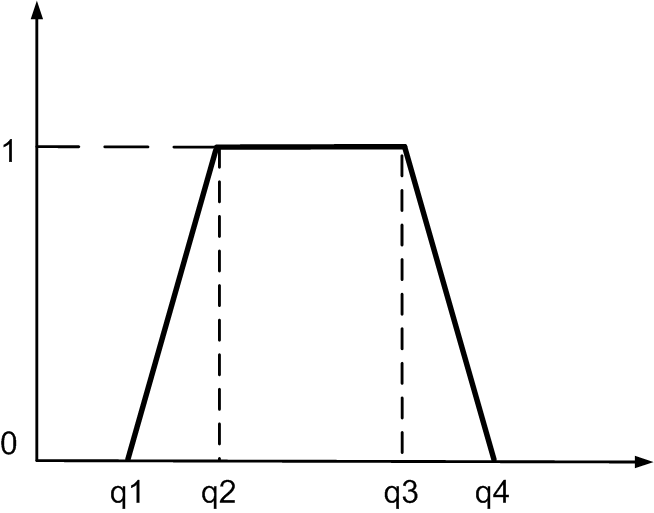} &
\includegraphics[height=2.2cm]{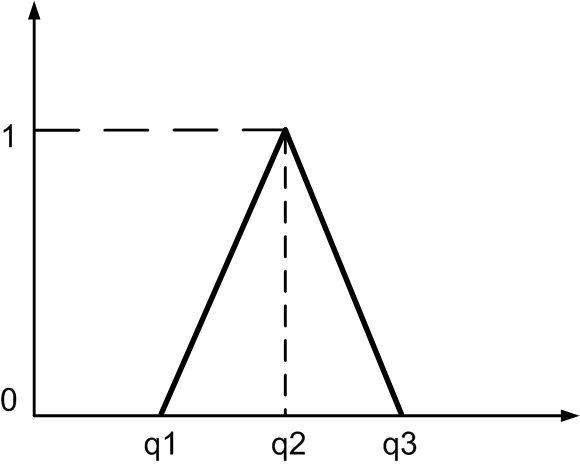} &
\includegraphics[height=2.2cm]{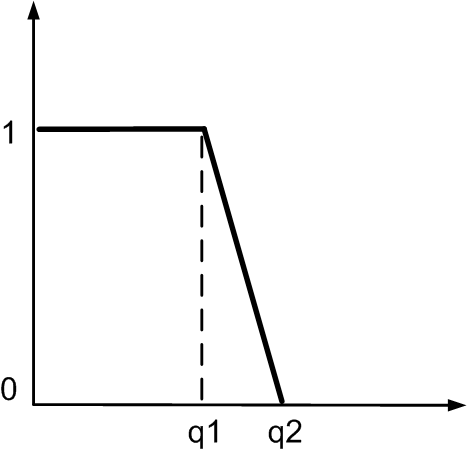} &
\includegraphics[height=2.2cm]{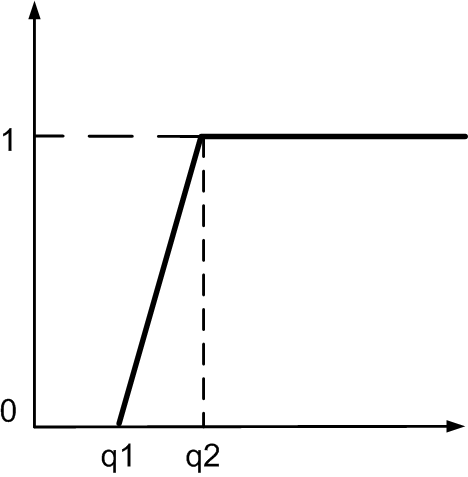} \\ 
(a) & (b) & (c) & (d) \\ 
\end{tabular}
\caption{(a) Trapezoidal ($\mathtt{trz}$); (b) Triangular ($\mathtt{tri}$); (c) Left-shoulder ($\mathtt{left}$); (d) Right shoulder ($\mathtt{right}$).} 
\label{fig:muf}
\end{center}
\end{figure*}

\nd To build fuzzy membership functions, common options are the trapezoidal (Figure~\ref{fig:muf} a), triangular (Figure~\ref{fig:muf} b), the left-shoulder (Figure~\ref{fig:muf} c), and the right-shoulder (Figure~\ref{fig:muf} d) functions. For example, the fuzzy set of \textsf{SecondAgeHuman} can be defined as $\mathtt{trz}(15, 18, 60, 70)$. If a person is $17$ years old, the membership degree to \textsf{SecondAgeHuman} can be evaluated 
as follows: 
\begin{equation}
\mu_{\textsf{SecondAgeHuman}}(\textsf{p17}) =
\mathtt{trz}(15, 18, 60, 70)(17) = 0.67 \ .
\end{equation}
Given a fuzzy set $A$ and a degree $\alpha \in [0,1]$, the \emph{$\alpha$-cut} of $A$, denoted $A_{\alpha}$, is the (crisp) set of elements of the domain of universe that have a membership degree to $A$ greater or equal than $\alpha$. That is:
\begin{equation}
A_{\alpha} = \{x \in X : \mu_{A}(x) \geq \alpha \} \ .
\end{equation}

\noindent The \emph{support} of a fuzzy set $A$ is the set of elements of $X$ that have a membership degree strictly greater than $0$, i.e.
\begin{equation}
support(A) = \{ u \in X : \mu_{A}(u) > 0 \} \ .
\end{equation}






\noindent Logical operations over classical
sets are also generalized to the fuzzy case. To compute the conjunction,
disjunction and complement over fuzzy sets one typically uses 
families of functions, called \emph{t-norm} function ($\otimes$) for set intersection, \emph{t-conorm} function ($\oplus$) for set union, and a \emph{negation} function ($\ominus$) for set complement (see, e.g.~\cite{Klir95}). For instance, the \emph{minimum} and the \emph{product} are t-norms, and, thus, set intersection is computed as  $\mu_{A\cap B}(x) = \min(\mu_A(x), \mu_B(x))$ and  
$\mu_{A\cap B}(x) = \mu_A(x) \cdot \mu_B(x)$, respectively.
The \emph{maximum}  and the \emph{algebraic sum} are t-conorms and, thus, 
set union is computed as 
$\mu_{A\cup B}(x) = \max(\mu_A(x), \mu_B(x))$ and
$\mu_{A\cup B}(x) = \mu_A(x) + \mu_B(x) - \mu_A(x) \cdot \mu_B(x)$,
respectively. The complement to one is a negation function and, thus, 
set complement is computed as $\mu_{\bar{A}}(x) = \mu_{X\setminus A}(x) =  1- \mu_A(x)$. 

Another important family of fuzzy functions are \emph{Aggregation Operators} (AOs)~\cite{Torra}. AOs have been widely used in computational intelligence because of their ability to fuse imprecise pieces of information. In this paper, \emph{Ordered Weighted Averaging} (OWA) operators~\cite{OWA} will be relevant. An OWA operator of dimension $n$ is an AO aggregating $n$ values $x_1, \dots, x_n \in [0, 1]$ such that:
\begin{equation}
\label{eq:owa}
\mathbf{OWA}_{[w_1, \dots, w_n]} ([x_1, \dots, x_n]) = \sum^n_{i=1} w_i b_i \ ,
\end{equation}

\noindent where $b_i$ is the $i$-th largest value of the $x_i$ for each $i \in [1, n]$. 
%
A possible way to learn the weights of an OWA operator is via a \emph{Regular Increasing Monotone} (RIM) quantifier $Q$~\cite{RIMs}, which satisfies the boundary conditions of AOs, namely $Q(0) = 0$ and $Q(1) = 1$, and is monotonic increasing, i.e., $Q(x_{1}) \leq Q(x_{2})$ when $x_{1} \leq x_{2}$. Given a RIM $Q$, the weights of an OWA weighting vector of dimension $n$ can be computed as ($i \in \{1, ..., n\}$)
\begin{equation}
\label{eq:qOWA}
w_i = Q(\frac{i}{n}) - Q(\frac{i-1}{n}) \ .
\end{equation}

\subsection{Fuzzy quantification} \label{sec:22}

\nd In the literature, there are two main types of \emph{fuzzy quantified sentences} (see Example~\ref{ex:sentences} below): namely,
\begin{description}
    \item[Type I sentences:] $Q$ of $X$ are $G$;
    \item[Type II sentences:] $Q$ of $F$ are $G$,
\end{description}
\nd where $Q$ is a fuzzy quantifier, $X$ is a crisp finite set, and $F$ and $G$ are two fuzzy subsets of $X$. 

\begin{example}
\label{ex:sentences}
Let us consider the following quantified sentences that will be used in the rest of this paper:
\begin{itemize}
    \item Type I: Few hotels are cheap;
    
    \item Type II: Most close hotels are cheap.  
\end{itemize}
\nd In the first quantified sentence, the fuzzy quantifier $Q$ is `few' and the fuzzy set $G$ is `cheap hotels'. In the second sentence, the fuzzy quantifier $Q$ is `most', the fuzzy set $F$ is `close hotels', while the fuzzy set $G$ is again `cheap hotels'.
\qed 
\end{example}


\nd Concerning fuzzy quantifiers, there are two types of fuzzy quantifiers: \emph{absolute quantifiers} $Q_a : \mathbb{R}^+_0 \to \unit$ (such as ``few'' or ``approximately three'') and \emph{relative quantifiers} $Q_r : \unit \to \unit$ (such as ``most'' or ``almost all''). 
%
%
%
Both types of sentences can employ absolute and relative sentences. However, following~\cite{Liu}, without loss of generality, it is possible to restrict to Type I sentences with absolute quantifiers and Type II sentences with relative quantifiers only. In particular:
\begin{itemize}
\item the evaluation of a Type I sentence 
\begin{quote}
$Q_r$ of $X$ are $G$ 
\end{quote}
\nd  with a relative quantifier $Q_r$ is the same as the evaluation of the Type I sentence 
\begin{quote}
$Q_a$ of $X$ are $G$,
\end{quote}
\nd where $Q_a(i) = Q_r(\frac{i}{\mid X \mid})$ is an absolute quantifier;

\item the evaluation of a Type II sentence 
\begin{quote}
$Q_a$ of $F$ are $G$
\end{quote}
\nd with an absolute quantifier $Q_a$ is equivalent to the evaluation of the Type I sentence 
\begin{quote}
$Q_a$ of $X$ are $F \cap G$.
\end{quote}

\end{itemize}

\nd Many different methods to evaluate Type I sentences and Type II sentences have been proposed in the literature~\cite{DelgadoRSV}. They are strongly based on the choice of the definition of \emph{cardinality} of a fuzzy set~\cite{DelgadoRSV}.
We next illustrate some proposals for the evaluation of quantified sentences.

\paragraph{Zadeh's method~\cite{ZadehQuantification}} 
The evaluation of a Type I sentence of the form `$Q$ of $X$ are $G$',
given a crisp set $X = \{x_1, \dots, x_n \}$, a fuzzy quantifier $Q$ and a fuzzy set $G$, is given by:
\begin{equation}
\label{eq:zadehTypeI}
\texttt{solveTypeI}(Q, X, G) = Q(card(G)) = Q \Bigg( \sum^n_{i=1} \mu_G(x_i)  \Bigg) \ ,
\end{equation}

%
%
%

%
%

\nd where the cardinality of a fuzzy set is the so-called \emph{sigma count}~\cite{Kosko86}; 
\begin{equation}
    card(G) = \sum^n_{i=1} \mu_G(x_i) \ .
\end{equation}

\begin{example}
\label{ex:hotel1}
Assume a knowledge base with three hotels with the following prices and distances to the city center:

\begin{center}
\begin{tabular}{|ccc|}
  \hline
  Hotel & Price  &  Distance \\
  \hline
  $h_1$ & $100$ & $900$ \\
  $h_2$ & $80$ & $100$ \\
  $h_3$ & $70$ & $400$ \\
  \hline
\end{tabular}

\end{center}

\nd Let us now consider the Type I sentence 
\begin{quote}
   Few hotels have a low price 
\end{quote}

 \nd (similar to the Type I sentence in Example~\ref{ex:sentences}), where the fuzzy quantifier ``few'' is defined by 
 \begin{equation*}
     Q_{\textsf{few}}(x) = \mathtt{left}(1, 3)(x)
 \end{equation*}

 \nd and the fuzzy set ``low price'' is defined as\footnote{Prices are defined in euros.} 
 \begin{equation*}
\mu_{\textsf{LowPrice}}(x) = \mathtt{left}(60,100)(x) \ .
 \end{equation*}
 
 \nd Therefore, for $X = Hotel = \{h_1, h_2, h_3 \}$ and $G = LowPrice = \{ 0.0/h_1, 0.5/h_2, 0.75/h_3\}$, using Zadeh's method, we have that 
 \begin{equation*}
     \emph{\texttt{solveTypeI}}(Q, X, G) = Q \Bigg( \sum^3_{i=1} \mu_G(h_i) \Bigg) = Q(0.0 + 0.5 + 0.75) = Q(1.25) = 0.875 \ .
 \end{equation*}
\qed
\end{example}
\nd On the other hand, the evaluation of a Type II sentences is given by:
\begin{equation} \label{eq:zadehTypeII}
\texttt{solveTypeII}(Q, F, G) =
Q \Bigg( \frac{ cardinality(F \cap G)}{cardinality(G)} \Bigg) = 
Q \Bigg( \frac{\sum^n_{i=1} \mu_{F \cap G}(x_i)} {\sum^n_{i=1} \mu_F(x_i)} \Bigg) \ .
\end{equation}

\begin{example}[Example~\ref{ex:hotel1} cont.]
\label{ex:hotel2}
Consider the Type II sentence 
\begin{quote}
    Most hotels with a low distance have a low price \ ,
\end{quote}

 \nd which is similar to the one in Example~\ref{ex:sentences}, where now the fuzzy set ``low distance'' is defined as\footnote{Distances are measured in meters.} 
 \begin{equation*}
  \mu_{\textsf{LowDistanceHotel}}(x) = \mathtt{left}(100,1100)(x)
 \end{equation*}

 \nd and the fuzzy quantifier ``most'' is defined as
 \begin{equation*}
     Q_{\textsf{most}}(x) = \mathtt{right}(0.5, 1)(x)  \ .
 \end{equation*}
\nd Now, as $F = LowDistanceHotel = \{ 0.2 / h_1, 1 / h_2, 0.7 / h_3\}$, and 
$F \cap G = \{ 0.5 / h_2, 0.7 / h_3\}$, we have that 
\begin{equation*}
\begin{array}{rcl}
    \emph{\texttt{solveTypeII}}(Q, F, G) & = & 
Q \Bigg( \frac{\sum^n_{i=1} \mu_{F \cap G}(x_i)} {\sum^n_{i=1} \mu_F(x_i)} \Bigg) \\
& = & Q \Bigg( \frac{0.5 + 0.7}{0.2 + 1 + 0.7} \Bigg) \\
& = &  Q( \frac{1.2}{1.9} ) = Q(0.6316) = 0.4386 \ .
\end{array}
 \end{equation*}
\qed
\end{example}

\paragraph{Yager's OWA-based method~\cite{OWA}} 

This method assumes a relative RIM quantifier $Q$. Specifically, given a vector of weights $[w_1, \dots, w_n]$ computed using Equation~\ref{eq:qOWA}, where $n = |support(G)|$, the evaluation of a Type I sentence is given by:
\begin{equation} \label{eq:yagerTypeI}
\texttt{solveTypeI}(Q, X, G) = \mathbf{OWA}_{[w_1, \dots, w_K]} (b_1, \dots, b_n) \ ,
\end{equation}

\noindent where $b_i$ is $i$-th largest value of $G$ for each $i \in [1, n]$. 

On the other hand, the evaluation of Type II sentences is given by:
\begin{equation}
\label{eq:yagerTypeII}
\texttt{solveTypeII}(Q, F, G) =  \sum_{i=1}^n w_i \cdot \beta_i \ ,
\end{equation}
\nd where $w_i = Q(S_i) - Q(S_{i-1})$, 
$S_i=\frac{\sum^i_{j=1} f_j}{\sum^n_{k=1} f_k}$,
$f_i$ is the $i$-th smallest $F(x_i)$, and
$\beta_i$ is the largest value of membership to the fuzzy set $\mathcal{B}(x_i) = \max(1 - F(x_i), G(x_i)), \forall i \in \{ 1, \dots , n \}$. 

\paragraph{GD method~\cite{GD}} 
This method is based on the ED fuzzy cardinality to evaluate Type I sentences.\footnote{GD and ED are not acronyms: they are the most important examples of the family $G$ of quantification methods and of the family $E$ of fuzzy cardinalities, respectively~\cite{DelgadoRSV}.} Specifically, the \emph{ED fuzzy cardinality} of a fuzzy set $G$ is a fuzzy set defined for each $k \in [0, n]$, where $n = |support(G)|$, as follows:
\begin{equation}
\label{eq:ED}
ED(G, k) = b_k - b_{k+1} \ ,
\end{equation}

\nd where $b_i$ is the $i$-th largest value of $G$, $b_0 = 1$, and $b_{n+1} = 0$.
Then, the GD method to evaluate Type I sentences is given by:
\begin{equation}
\label{eq:GD}
\texttt{solveTypeI}(Q, X, G) = 
\sum_{i=0}^n ED(G, i) \cdot Q(i) \ .
\end{equation}


\begin{example}
Let us evaluate Example~\ref{ex:hotel1} using GD. Let us observe that
\[
b_0 = 1, b_1 = 0.75, b_2 = 0.5, b_3 = 0 \ .
\]
\nd Next, let us now compute each value of the summation in Equation~\ref{eq:GD}. 
\begin{description}

    \item[i=0:] $ED(G, 0) \cdot Q(0) = (b_0 - b_1) \cdot Q(0) = (1 - 0.75) \cdot Q(0) = 0.25 \cdot 1 = 0.25$ 

    \item[i=1:] $ED(G, 1) \cdot Q(1) = (b_1 - b_2) \cdot Q(1) = (0.75 - 0.5) \cdot Q(1) = 0.25 \cdot 1 = 0.25$

    \item[i=2:] $ED(G, 2) \cdot Q(2) = (b_2 - b_3) \cdot Q(2) = (0.5 - 0) \cdot Q(2) = 0.5 \cdot 0.5 = 0.25$. 
    
\end{description}
\nd As a consequence, 
\[
\texttt{\emph{solveTypeI}}(Q, X, G) = \sum_{i=0}^2 ED(G, i) \cdot Q(i) = 0.25 + 0.25 + 0.25 = 0.75 \ . 
\]
\qed
\end{example}
\nd The work~\cite{GD} also generalized the method to work on Type II sentences, which we illustrate next. 

We may assume, without loss of generalization, that $F$ is a normal fuzzy set.
%
For convenience, it is useful to introduce some auxiliary value definitions. 
The \emph{relative cardinal} of $G$ with respect to $F$ and $\alpha \in [0,1]$ is given by:
\begin{equation}
\label{eq:C}
C(G/F, \alpha) = \frac{|(G \cap F)_{\alpha}|}{|F_{\alpha}|} \ .
\end{equation}
%
%
%
%
The \emph{set of relevant $\alpha$-cuts} of a set $A$ is given by:
\begin{equation}
\label{eq:cuts}
cuts(A) = \{ \alpha \in (0, 1] \mid \exists x_i\in A \textrm{ such that } \mu_A(x_i) = \alpha \} \ . 
\end{equation}
\nd Let us denote $\{ \alpha_0, \alpha_1, \dots, \alpha_p \}$ as the reordering of the set $cuts(G \cap F) \cup cuts(F) \cup \{ 0 \}$ such that $\alpha_i > \alpha_{i+1}$ for each $i \in \{0, p-1\}$. Note that $\alpha_0 = 1$ if $F$ is normal and $\alpha_p = 0$.
%
%


Finally, following~\cite[Property 6.1.3]{GD}, the \emph{GD method} to evaluate a Type II sentence is defined as:
%
\begin{equation}
\label{eq:GD2alt}
\texttt{solveTypeII}(Q, F, G) = 
\sum_{\alpha_i \in cuts(G \cap F) \cup cuts(F)} (\alpha_i - \alpha_{i+1})  \cdot Q(C(G/F, \alpha_i)) \ .
\end{equation}


\begin{example}
Let us evaluate now Example~\ref{ex:hotel2} using GD. 

It can be verified that 
\[
cuts(G \cap F) \cup cuts(F) = \{ 0.5, 0.7, 0.2, 1 \} \ . 
\]
\nd and that
\[
\alpha_0 = 1, \alpha_1 = 0.7, \alpha_2 = 0.5, \alpha_3 = 0.2, \alpha_4 = 0 \ .
\]
\nd Consequently, we have 
\begin{eqnarray*}
C(G/F, 1) & = & \frac{|(G \cap F)_{1}|}{|F_{1}|} = \frac{|\emptyset|}{|\{ h_2 \}|} = 0    \\
C(G/F, 0.7) & = & \frac{|(G \cap F)_{0.7}|}{|F_{0.7}|} = \frac{|\{ h_3 \}|}{|\{ h_2, h_3 \}|} = 0.5 \\
C(G/F, 0.5) & = & \frac{|(G \cap F)_{0.5}|}{|F_{0.5}|} = \frac{|\{ h_2, h_3 \}|}{|\{ h_2, h_3 \}|} = 1 \\
C(G/F, 0.2) & = & \frac{|(G \cap F)_{0.2}|}{|F_{0.2}|} = \frac{|\{ h_2, h_3 \}|}{|\{ h_1, h_2, h_3 \}|} = 0.67 \ .
\end{eqnarray*}
\nd Therefore, 
\begin{eqnarray*}
\texttt{\emph{solveTypeII}}(Q, F, G) & =  &
(1 - 0.7) \cdot Q(C(G/F, 1)) + (0.7 - 0.5) \cdot Q(C(G/F, 0.7)) \\
&& + \ (0.5 - 0.2) \cdot Q(C(G/F, 0.5))  + (0.2 - 0) \cdot Q(C(G/F, 0.2)) \\
& = &  0.3 \cdot Q(0) + 0.2 \cdot Q(0.5) + 0.3 \cdot Q(1)  + 0.2 \cdot Q(0.67) \\
& = & 0.3 \cdot 0 + 0.2 \cdot 0 + 0.3 \cdot 1 + 0.2 \cdot 0.34 = 0.368 \ . 
\end{eqnarray*}
\qed
\end{example}


\subsection{Semantic Web technologies}
\label{sec:23}

\nd This section recalls some basic notions within the  Semantic Web domain that will be used in the rest of the paper.

\subsubsection{Ontologies}
\label{sec:ontologies}

\nd An ontology is a formal and shared specification of the knowledge in a domain~\cite{OntologyHandbook}. An ontology offers a common vocabulary, terminology, a conceptual model, and reasoning capabilities. \emph{OWL} (Web Ontology Language) is the recommendation of the World Wide Web Consortium (W3C) to represent ontologies. Its semantics is based on a Description Logics and the current version is OWL~2~\cite{OWL2W3C,OWL2JWS}. 

The main ingredients of OWL ontologies to model a domain are \emph{concepts/classes} (e.g., the class of hotels \textsf{Hotel}), \emph{instances/individuals} (e.g., the hotel \textsf{h1}), \emph{properties} (binary relationships between a pair of elements), and \emph{datatypes} (elements that do not belong to the represented domain, but to a domain already known to the machine, e.g., numbers or strings). There are two types of properties: \emph{object properties}, such as \textsf{hasAlbum}, relate a pair of individuals (a music artist/band and an album), whereas \emph{data properties}, such as \textsf{price} or \textsf{distance}, link an individual with a data value of a datatype. Complex classes can be built from previously defined ontology elements.

\begin{example}
To evaluate the two first queries of Example~\ref{ex:blackMetal}, we can consider the classes $\textsf{Band} \sqcap \exists \textsf{hasAlbum}.\textsf{BlackMetal}$ and $\textsf{Band} \sqcap \forall \textsf{hasAlbum}.\textsf{BlackMetal}$ (in DL syntax).\footnote{Technically,
$\textsf{Band} \sqcap \exists \textsf{hasAlbum}.\textsf{BlackMetal}$ is the set of objects $o$ that belong to the \textsf{Band} class and to the $\exists \textsf{hasAlbum}.\textsf{BlackMetal}$ class, i.e., which are related via the binary relation \textsf{hasAlbum} to \emph{some} object $o'$ being an instance of the \textsf{BlackMetal} class. Moreover, $\textsf{Band} \sqcap \forall \textsf{hasAlbum}.\textsf{BlackMetal}$ is the class of objects $o$ that belong to the \textsf{Band} class and such that \emph{all} objects $o'$ related to $o$ via \textsf{hasAlbum} are instances of the \textsf{BlackMetal} class.}.  \qed
\end{example}
\nd The knowledge in an ontology is organized in \emph{axioms} (constraints or restrictions to be satisfied by the elements of the ontology) such as \emph{concept assertions} (stating that an individual is an instance of a concept/class), \emph{data property assertions} (linking an individual and a datatype value via a data property), or \emph{subclass axioms} (stating that a class is more specific than another one).  

\begin{example}[Example~\ref{ex:hotel1} cont.]
The table can be encoded via the following ontology (in DL syntax), with $3$ class assertions involving the \textsf{Hotel} class and $6$ data property assertions
involving the \textsf{price} and \textsf{distance}data properties:
:
{\footnotesize
\begin{eqnarray*}
\ \O & \!\!\!\!\!\! =\{ &  \textsf{Hotel}(h_1), \textsf{price}(h_1, 100), \textsf{distance}(h_1, 900), \\
& & \textsf{Hotel}(h_2), \textsf{price}(h_2, 80), \textsf{distance}(h_2, 100), \\
& & \textsf{Hotel}(h_3), \textsf{price}(h_3, 70), \textsf{distance}(h_3, 400) \ \ \} \ .  
\end{eqnarray*}
}
\qed
\end{example}

\nd Reasoning is one of the most distinguishing features related to ontologies, playing an important role both in the development of models (to verify their correctness) and in the deployment of real-world applications (to discover knowledge). A {\emph{reasoner}~\cite{Khamparia2017} is a software implementation of algorithms that solve some reasoning tasks}. In other words, it infers implicit knowledge from explicit knowledge in a logically consistent way. Several semantic reasoners support OWL~2 ontologies and provide different reasoning tasks, such as 
classification (i.e., computing a concept and property hierarchy based on inferred subclass relationships) or instance retrieval (i.e., retrieving all the instances of a given class)~\cite{DLtextbook}. 

Fuzzy ontologies are extensions of ontologies with elements from fuzzy logic and fuzzy set theory~\cite{FuzzyOntologies}. For example, fuzzy concepts denote fuzzy sets of individuals, and fuzzy datatypes can be represented using fuzzy membership functions (such as those in Figure~\ref{fig:muf}). Of course, the fuzzy case requires different reasoning algorithms and the definitions of the reasoning tasks are generalized as well. For example, the instance retrieval task now involves retrieving graded instances, i.e., not only the instances of a given class but also their membership degrees to the class. Popular fuzzy ontology reasoners are fuzzyDL~\cite{KBS2016} and DeLorean~\cite{ESWA2012}, whereas the most popular fuzzy ontology language is Fuzzy OWL~2~\cite{IJAR2011}.

\subsubsection{Knowledge graphs}
\label{sec:kgs}

\nd A \emph{Knowledge Graph} (KG) is ``a graph of data intended to accumulate and convey knowledge of the real world, whose nodes represent entities of interest and whose edges represent relations between these entities''~\cite{kg-book}. Classical examples of knowledge graphs are \emph{DBPedia}~\cite{DBpedia} and Wikidata~\cite{Wikidata}. In practice, the majority of KGs are RDF graphs. 

\emph{RDF} (Resource Description Framework) is a standard formalism for data interchange~\cite{RDF} which uses a triple representation $\langle s, p, o \rangle$ where $s$ is a \emph{subject} (a resource), $o$ an \emph{object} (a resource or literal), and $p$ is a \emph{property/predicate}, stating that $s$ is related to $o$ via the property $p$. For example, the property \textsf{rdf:type} states that the triple subject belongs to a certain class (the triple object). There are several syntaxes to serialize RDF triples; in this paper we will use Turtle~\cite{RDFturtle}.

\begin{example}[Example~\ref{ex:hotel1} cont.]
The table can be encoded via the following set of RDF triples (in Turtle syntax):

\begin{lstlisting}[
    frame=single,
    numbers=left,
    basicstyle=\scriptsize\ttfamily,
    breaklines=true
]
@PREFIX rdf: <http://www.w3.org/1999/02/22-rdf-syntax-ns#> .
:h1 rdf:type :Hotel .
:h1 :price 100 .
:h1 :distance 900 .
:h2 rdf:type :Hotel .	
:h2 :price 80  .
:h2 :distance 100 .
:h3 rdf:type :Hotel .
:h3 :price 70 .
:h3 :distance 400 .
\end{lstlisting}
%
\qed
\end{example}
\nd To define the schema that an RDF graph follows, different languages with different expressivities can be used, ranging from RDF-Schema, also called RDFS~\cite{RDFS}, to OWL 2~\cite{OWL2JWS}. For example, property \textsf{rdfs:subclassOf} states that a class (the triple subject) is a sublass of another one (the triple object). While using RDFS one can only define hierarchies of concepts and properties, domain and ranges of properties, and the classes that individuals belong to, OWL~2 provides a much richer language (see previous section)~\cite{DLtextbook}.

SPARQL is RDF standard query language~\cite{SPARQL,SPARQL2}. Among other things, it is used to retrieve triples from a data graph that satisfy given conditions. A SPARQL query is submitted to a SPARQL \emph{endpoint} for evaluation (a server capable of processing SPARQL queries). The KGs \emph{DBPedia} and Wikidata  can be accessed/queried through SPARQL \textit{endpoints}.\footnote{\url{https://dbpedia.org/sparql} and \url{http://query.wikidata.org}, respectively.}

In the literature, there are also fuzzy extensions of KGs. In particular, \emph{fuzzy RDF graphs} are based on fuzzy RDF statements of the form $\langle s, p, o, \alpha \rangle$, where $\alpha \in [0,1]$~\cite{Straccia09f}. The Fuzzy RDF system, based on a fuzzy extension of $\rho$df, makes it possible to answer queries over local fuzzy RDF graphs~\cite{Straccia09f}. However, there are currently no public endpoints supporting such fuzzy statements.

\section{Fuzzy Quantification over OWL ontologies and KGs}
\label{sec:queries}

\nd One of the key concepts behind our approach is the fact that, even if the ontology only contains non-fuzzy information, given a functional numerical data property $d$, such as the price of a hotel room, it is possible to consider a fuzzy linguistic label $L$ (e.g., \emph{low} price), represented using a fuzzy datatype, to build a fuzzy concept of the form $\exists d.L$. 

\begin{example}
Given a classical ontology about accommodation types, given a datatype property \textsf{hasPrice}, associating to each hotel the price of a room, we can consider the fuzzy linguistic label \textsf{LowPrice} defined as left shoulder function over the crisp set $X$ of hotel prices and build the fuzzy concept 
\[
G = \textsf{Hotel} \sqcap \exists \textsf{hasPrice}.\textsf{LowPrice} \ .
\]
\nd To `retrieve all hotels whose price is low', one can solve the fuzzy instance retrieval problem over $G$.

\qed
\end{example}

\nd In this section, we will firstly discuss how to obtain and represent such fuzzy linguistic labels and then we will propose different types of quantification queries over semantic knowledge bases. Finally, we will discuss how to represent the queries.

\subsection{Representing and computing the fuzzy linguistic labels in ontologies}
\label{sec:31}

\begin{figure}[!t]
\begin{center}
\includegraphics[scale=0.5]{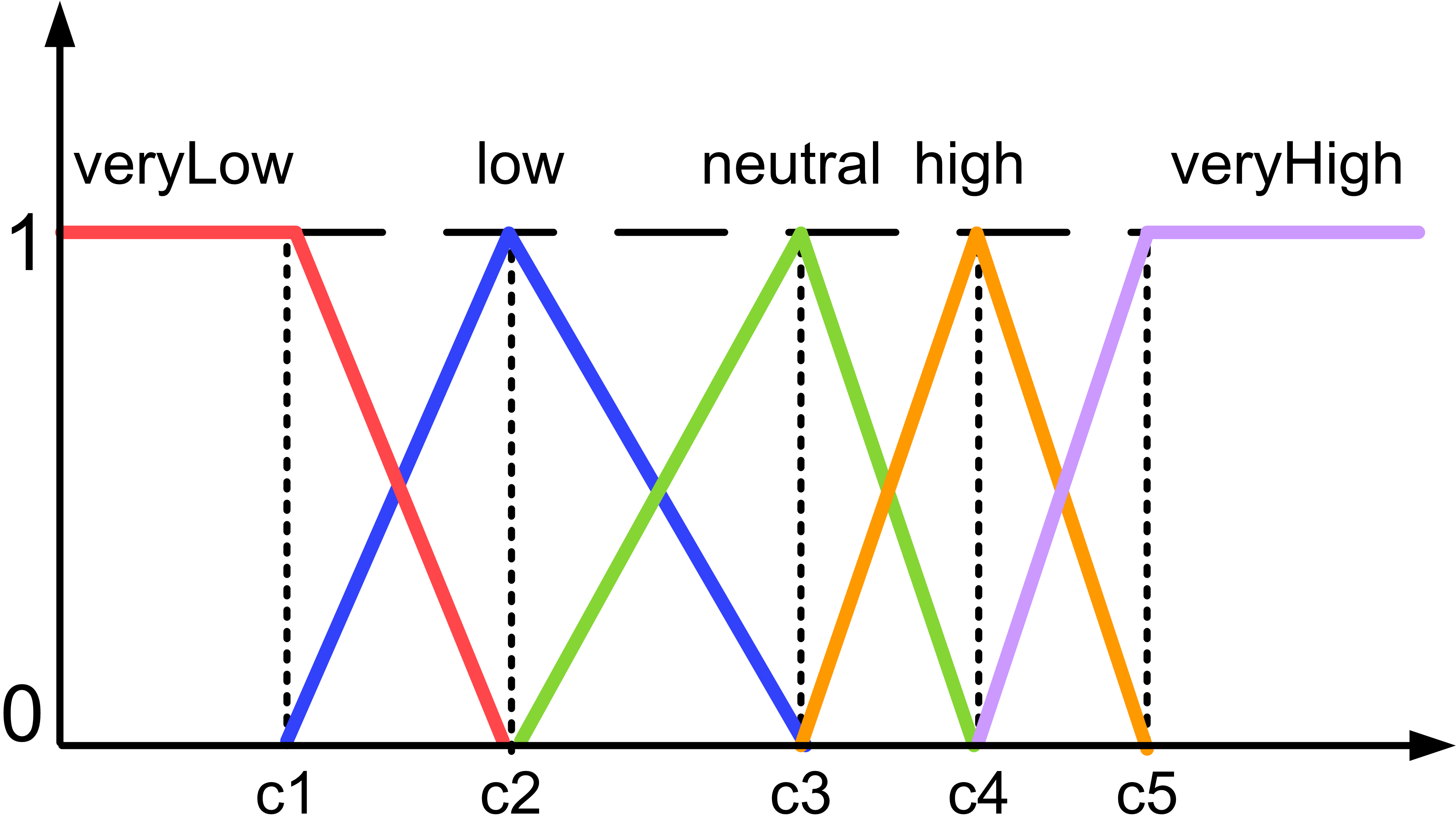}
\end{center}
\caption{Fuzzy sets over centroids, learned from data via c-means clustering algorithm.} \label{fig:muf2}
\end{figure}

\nd Fuzzy linguistic labels can be represented using Fuzzy OWL~2 datatypes~\cite{IJAR2011}, which includes trapezoidal, triangular, left-shoulder, and right-shoulder fuzzy membership functions. Fuzzy OWL~2 is a language for fuzzy ontology representation which uses standard OWL~2 files but encodes the fuzzy knowledge using OWL~2 annotations. In particular, a fuzzy datatype is represented by annotating an OWL~2 datatype, using an XML-like syntax to specify the type and parameters of the fuzzy membership function. For example, the following annotation specifies a left-shoulder function with parameters $60$ and $100$:

\begin{lstlisting}[
    frame=single,
    numbers=left,
    basicstyle=\scriptsize\ttfamily,
    breaklines=true
]
<fuzzyOwl2 fuzzyType=\"datatype\">
     <Datatype type=\"leftshoulder\" a=\"60\" b=\"100\" />
</fuzzyOwl2>
\end{lstlisting}


\nd It could also be the case that the ontology does not include fuzzy datatypes. This happens, for instance, if we want to query well known ontologies (such as the Music ontology\footnote{\url{http://purl.org/ontology/mo/}}). 
In such cases, it has been shown in the literature how to automatically compute the linguistic labels corresponding to a functional numerical data property by learning them from examples included in a classical ontology (see, e.g.~\cite{Cardillo24,Cardillo22,Datil,Lisi15}). Typically, they employ a centroid-based clustering algorithm and then use the centroids to automatically build membership functions 
(as illustrated in Figure~\ref{fig:muf2}). 
 Usually, one learns the membership functions by taking into account all values of a  data property. In other cases, we would be interested to restrict only a subset of it, e.g., to the prices of hotel. In this way, for example, the lower prices of hostels affect the prices of hotels if hostels are represented as subclasses of hotels, but they do not affect them if  hostels are represented as a disjoint class to hotels (but both are subclasses of the accommodation class). 
 
 In general, given the data property $d$, we will compute the list of $d$ range values, store them (e.g., in a column of a CSV file) and apply a membership computation algorithm to it~\cite{Cardillo24,Cardillo22,Lisi15} (see Equation~\ref{eq:G} later on). In particular, there is an implementation called Datil~\cite{Datil} for this purpose.

\subsection{Representing and computing the fuzzy linguistic labels in knowledge graphs}

\nd Interestingly, the previous annotations to characterize fuzzy datatypes can also be encoded as triples in a knowledge graph~\cite{TEL}. For example:

\begin{lstlisting}[
    frame=single,
    numbers=left,
    basicstyle=\scriptsize\ttfamily,
    breaklines=true
]
@PREFIX ex: <http://www.example.org/> .
@PREFIX owl: <http://www.w3.org/2002/07/owl#> .
@PREFIX rdf: <http://www.w3.org/1999/02/22-rdf-syntax-ns#> .
@PREFIX rdfs: <http://www.w3.org/2000/01/rdf-schema#> .
ex:fuzzyLabel rdf:type owl:AnnotationProperty .
ex:LowPrice rdf:type rdfs:Datatype .
ex:LowPrice ex:fuzzyLabel """<fuzzyOwl2 fuzzyType=\"datatype\">
 <Datatype type=\"leftshoulder\" a=\"60\" b=\"100\" />
 </fuzzyOwl2>""" .
\end{lstlisting}


\nd Therefore, fuzzy datatypes and their definitions can also be retrieved using a SPARQL query such as:
\begin{lstlisting}[
    frame=single,
    numbers=left,
    basicstyle=\scriptsize\ttfamily,
    breaklines=true
]
PREFIX rdf: <http://www.w3.org/1999/02/22-rdf-syntax-ns#> .
SELECT ?x ?y
WHERE {
   ?x rdf:type rdfs:Datatype .
   ?x ex:fuzzyLabel ?y
}
\end{lstlisting}
%
%
\nd As in the previous OWL ontology case, it is possible that the knowledge graph does not include fuzzy datatypes (for example, if we query DBpedia). Interestingly, a similar solution is possible with knowledge graphs: one can retrieve the values of a data property from a data property using a SPARQL query, store them, 
and use the same algorithms as mentioned at the end of  Section~\ref{sec:31} to learn linguistic labels from them. An example of a query is:

\begin{lstlisting}[
    frame=single,
    numbers=left,
    basicstyle=\scriptsize\ttfamily,
    breaklines=true
]
PREFIX rdf: <http://www.w3.org/1999/02/22-rdf-syntax-ns#> .
SELECT ?y
WHERE {
   ?x rdf:type :C .
   ?x :d ?y 
}
\end{lstlisting}


\nd whose aim is to retrieve all $d$-values of instances of the class $C$.

\subsection{Type I sentences over classical knowledge bases}
\label{sec:case1}

\nd Let us recall the previous observation that, given a functional numerical data property $d$, one can build a fuzzy concept of the form $\exists d.L$ using a fuzzy linguistic label $L$, represented using a fuzzy datatype. Therefore, given a classical (i.e., crisp) semantic knowledge base $\S$, a fuzzy quantifier $Q$, a crisp class 
$C$, a functional numerical data property $d$, and a fuzzy linguistic label $L$, Type I sentences are in this case  of the following form:

\begin{description}
    \item[Type I sentences:] $Q$ of $C$ have a value of property $d$ which is $L$.
\end{description}

\begin{example}
In the Type I sentence ``Few hotels have a low price'', ``few'' is a fuzzy quantifier, ``Hotel'' is a class, ``price'' is a numerical data property, and ``low'' is a fuzzy datatype. \qed
\end{example}

\noindent Clearly, there are four parameters, $Q$, $C$, $d$, and $L$. We will assume that $Q$ and $L$ are represented as Fuzzy OWL~2 datatypes. Furthermore, if $Q$ is relative, the domain of the fuzzy datatype should be $\unit$, e.g., 
$\mathtt{tri}(0.4, 0.7, 1)$. Note that $C$ does not need to be a so-called  \emph{atomic} class, i.e. a class/concept name. For example, it could be a conjunction or a union of classes.

If $\S$ is an OWL ontology $\O$, direct instances of $C$ are represented using OWL~2 class assertion axioms, and the value of a property $d$ for an individual can be represented using an OWL~2 data property assertion. However, it is very important to note that there can be indirect instances or property values, i.e., they are not explicitly represented as axioms but can be inferred from the axioms in the ontology.

On the other hand,  if $\S$ is a knowledge graph $\K$, the case is very similar. Now, direct instances of $C$ will be represented using an \textsf{rdf:type} triple, and the value $v$ of a property $d$ for an individual $i$ will be represented using an RDF triple $\langle i, d, v \rangle$. Again, there can be indirect instances or property values, and  we will assume that $Q$ and $L$ are represented as Fuzzy OWL~2 datatypes. That is, $Q$ and $L$ will be the subjects of a triple related via \textsf{fuzzyLabel} with a Fuzzy OWL~2 annotation.

\subsection{Type II sentences over classical knowledge bases}
\label{sec:case3}

\nd Given a fuzzy quantifier $Q$, a crisp concept $C$, two functional numerical data properties $d_1$ and $d_2$, and two fuzzy linguistic labels $L_1$ and $L_2$, 
Type II sentences are of the following form:
\begin{description}
    \item[Type II sentences:] $Q$ of the $C$ with a value of property $d_1$ which is $L_1$, have a value of property $d_2$ which is $L_2$.
\end{description}
\begin{example}
In the Type II sentence ``Most hotels with small room size have a low price'', ``Most'' is a fuzzy quantifier, ``Hotel'' is a class, ``size'' and ``price'' are numerical data properties, and ``small'' and ``low'' are fuzzy datatypes. \qed
\end{example}
\nd Now, there are six parameters, $Q$, $C$, $d_1$, $d_2$, $L_1$, and $L_2$. We will assume that $Q$, $L_1$, and $L_2$ are represented as Fuzzy OWL~2 datatypes. Again, if $Q$ is relative, the domain of the fuzzy datatype should be $\unit$, e.g., $\mathtt{tri}(0.4, 0.7, 1)$.

\subsection{Type II sentences over fuzzy semantic knowledge bases}

\nd For the sake of completeness, let us also consider the generalization of the cases in Sections~\ref{sec:case1} and~\ref{sec:case3} to fuzzy ontologies and fuzzy knowledge graphs. Given a fuzzy semantic knowledge base $\S$, 
a fuzzy quantifier $Q$, and fuzzy classes $F, G$, fuzzy quantified sentences are of the following form:
\begin{description}
\item[Type II sentences:] $Q$ of $F$ are $G$.
\end{description}
\nd Note that now the sentences have three parameters. A fuzzy class can include a fuzzy linguistic label in its definition (e.g., $\exists \textsf{price}.\textsf{LowPrice}$), but that is not necessarily mandatory (e.g., \textsf{CheapHotel} class). Note also that if $F$ is a crisp concept, then the above is a Type I sentence.

\subsection{On representing queries}

\nd In some cases, it could be interesting to represent both the domain knowledge and the fuzzy quantified queries using the same language. To do so, we will discuss in this section how to represent such queries using Fuzzy OWL~2, a de-facto standard for fuzzy ontology representation.

An option is to extend Fuzzy OWL~2 syntax to include a new constructor, specifying the query type and the values of the relevant parameters. For Type I queries, there are four parameters ($Q$, $C$, $d$, and $L$). Therefore, a possible approach is to annotate a named concept (representing the query) as follows:

\begin{lstlisting}[
    frame=single,
    numbers=left,
    basicstyle=\scriptsize\ttfamily,
    breaklines=true
]
<fuzzyOwl2 fuzzyType="concept">
  <Concept 
     type="quantificationI" 
     quantifier="<STRING>" 
     concept="<STRING>" 
     property="<STRING>" 
     datatype="<STRING>"
  />
</fuzzyOwl2>
\end{lstlisting}


\noindent where the values of the attributes \textsf{quantifier} and \textsf{datatypes} are the names of two fuzzy datatypes, already defined using Fuzzy OWL~2 syntax, and the values of the attributes \textsf{concept} and \textsf{property} are the names of a concept and a numerical data property in the ontology, respectively. For example:

\begin{lstlisting}[
    frame=single,
    numbers=left,
    basicstyle=\scriptsize\ttfamily,
    breaklines=true
]
<fuzzyOwl2 fuzzyType="concept">
  <Concept 
     type="quantificationI" 
     quantifier="Few" 
     concept="Hotel" 
     property="hasPrice" 
     datatype="LowPrice"
  />
</fuzzyOwl2>
\end{lstlisting}


\nd Note that we use the same attribute (\textsf{quantifier}) for both absolute and relative quantifiers. Indeed, from the definition (the domain) of the quantifier, one can see whether the quantifier is absolute or relative. 

Note also that we have created and annotated a new concept to represent this query. Indeed, we could think of these types of concepts as subclasses of a more general \textsf{Query} concept, denoting more general types of queries over an ontology.




\section{Evaluating fuzzy quantified sentences}
\label{sec:solving}

\subsection{Evaluating Type I sentences over classical knowledge bases}

\begin{algorithm}[htbp]
\caption{Evaluating Type I quantified queries over knowledge bases}
\label{alg:1}
\algorithmicrequire \ A knowledge base $\S$, a classical concept $C$, a functional numerical data property $d$, a fuzzy datatype $L$, and a fuzzy quantifier $Q$

\algorithmicensure \ $\alpha \in [0, 1]$ as the evaluation of `$Q$ of $C$ are $\exists d.L$' 
\begin{algorithmic}[1]


\STATE $\S \gets \texttt{preprocess}(\S)$ \label{code:start}  \label{code:classify}     


\STATE $X \gets \texttt{getInstances}(\S, C)$ \label{code:task1} 

\STATE $G \gets \emptyset$ \label{code:iniG}

\FORALL{$i \in X$}   

        \STATE $v_i \gets \texttt{getPropertyValue}(\S, i, d)$  \label{code:task2}

        \IF {$v_i \neq\textsc{null}$} 
            \STATE $\alpha_i \gets L(v_i)$
        \ELSE
            \STATE $\alpha_i \gets 0$
        \ENDIF
  
    \STATE $G \gets G \cup \{  \alpha_i/i \}$ \label{code:finG}

\ENDFOR

\RETURN $\texttt{solveTypeI}(Q, X, G)$ \label{code:method}

\end{algorithmic}
\end{algorithm}

\nd Algorithm~\ref{alg:1} describes the process to evaluate a Type I sentence over a  knowledge base. The algorithm is agnostic about the knowledge base type (an ontology or a KG). The preprocessing (Line~\ref{code:classify}), the instance retrieval problem (Line~\ref{code:task1}), and the retrieval of the values of a property for a given individual (Line~\ref{code:task2}) is addressed in a different way for an ontology and for a KG.
The algorithm is also agnostic about the method used to evaluate a Type I sentence `$Q$ of $C$ are $L$' (Line~\ref{code:method}). For example, one can consider Zadeh's method (Equation~\ref{eq:zadehTypeI}) or the GD method (Equation~\ref{eq:GD}).

Note that before calling that method, Lines~\ref{code:iniG}--\ref{code:finG} compute a fuzzy set 
\begin{equation}
\label{eq:G}
    G = \{ L(d(i_1)) / i_1, \dots, L(d(i_k)) / i_k \}     \ ,
\end{equation}
\noindent where $i_1, \dots, i_k$ are the instances of concept $C$ and $d(i)$ denotes the value of data property $d$ for the individual $i$. 
Please note that $d$ is a functional property, so the value for a given individual (Line~\ref{code:task2}) is either unique or undefined (\textsc{null}).
Another important observation is the fact that the methods to evaluate the Type I sentence `$Q$ of $C$ are $L$' assume fuzzy sets (in our case, fuzzy subsets of the set of instances of $C$) to have a finite number of elements.

\begin{example}
Let us consider a variant of Example~\ref{ex:hotel1} with more hotels. Assume an ontology has $9$ hotels (instances of the \textsf{Hotel} class), namely $h_1, \dots, h_9$ with the following prices (per night, in a fixed currency): $50$, $60$, $70$, $80$, $100$, $120$, $140$, $140$, and $155$, respectively. We want to evaluate the Type I sentence `few hotels have a low price', where `few' is defined as the function $\mathtt{left}(3, 4)$.

Firstly, we fuzzify the domain of price values into three fuzzy labels \emph{low}, \emph{medium}, and \emph{high} price. To do so, a clustering algorithm computes the following three centroids: $60$, $100$, and $145$, respectively. Thereafter, we define the linguistic label \textsf{LowPriceHotel} as the function $\mathtt{left}(60, 100)$. Rather than computing the cardinality of the infinite fuzzy set $\mathtt{left}(60, 100)$, which is a fuzzy set over the reals, we compute the cardinality of the fuzzy set
\[
\{ 1.0 / h_1, 1.0 / h_2, 0.75 / h_3, 0.5 / h_4 \} \ ,
\]
\noindent which is restricted to the instances of \textsf{Hotel} class. 

Then, using Zadeh's method to evaluate the Type I sentence (Equation~\ref{eq:zadehTypeI}), the result is $Q(1.0 + 1.0 + 0.75 + 0.5) = Q(3.25) = \mathtt{left}(3, 4)(3.25) = 0.75$. \qed
\end{example}

\nd Clearly, the previous algorithm can easily be generalized to the case of several properties and fuzzy datatypes, as shown in Algorithm~\ref{alg:2}. Note that it is not necessary to consider several input concepts, as $C$ could already be an intersection of concepts. The main difference is that the numerical degrees obtained for each property are combined using some t-norm (Line~\ref{alg:tnorm}). Since $0$ acts as the absorbing element of each t-norm, if the value of some of the data properties is unknown for a given individual, then such individual belongs to the fuzzy set $G$ with degree $0$ (Line~\ref{alg:11}).

\begin{algorithm}[htbp]
\caption{Generalization of Algorithm~\ref{alg:1} to several data properties.}
\label{alg:2}
\algorithmicrequire \ A semantic knowledge base $\S$, a crisp concept $C$, a list of functional numerical data properties $D = [d_1, \dots, d_n]$, a list of fuzzy datatypes $L = [L_1, \dots, L_n]$, and a fuzzy quantifier $Q$

\algorithmicensure \ $\alpha \in [0, 1]$ as the evaluation of `$Q$ of $C$ are $\exists d_1.L_1 \sqcap \cdots \sqcap \exists d_n.L_n$'
%
\begin{algorithmic}[1]


\STATE $S \gets \texttt{preprocess}(\S)$ 


\STATE $X \gets \texttt{getInstances}(\S, C)$ 

\STATE $G \gets \emptyset$

\FORALL{$i \in X$}   

    \STATE $\alpha_i \gets 1$

    \FORALL{data property $d_j$}  
    
        \STATE $v_{ij} \gets \texttt{getPropertyValue}(\S, i, d_j)$  

        \IF {$v_{ij} \neq \textsc{null}$} 
            \STATE $\alpha_i \gets \alpha_i \otimes L_j(v_{ij})$ \label{alg:tnorm}
        \ELSE
            \STATE $\alpha_i \gets 0$ \label{alg:11}
        \ENDIF
    \ENDFOR  
  
    \STATE $G \gets G \cup \{  \alpha_i/i \}$ 

\ENDFOR

\RETURN $\texttt{solveTypeI}(Q, X, G)$ 

\end{algorithmic}
\end{algorithm}
Next, we examine the specific cases of OWL ontologies and RDFS knowledge graphs.

\subsubsection{The case of OWL ontologies}

\nd In this case, preprocessing (Line~\ref{code:classify}) typically involves an optional classification of the ontology, which speeds future computations up.

Line~\ref{code:task1} consists of solving the instance retrieval problem for the concept $C$ (that is, retrieving all instances of $C$), i.e.,
\[
X = \{i \in \O \mid \O \models C(i) \} \ .
\]
\nd Line~\ref{code:task2} retrieves the $d$-value of an individual $i$, i.e., 
\[
\{v \in \mathcal{O} \mid \O \models d(i, v) \} \ .\footnote{Or, equivalently,  
$\{v \in \mathcal{O} \mid \O \models (\exists d.=_v)(i) \}$.}
\]
\nd All these tasks (ontology classification, instance retrieval problem, and the retrieval of the values of a property for a given individual) can be solved by using current ontology reasoners. 

\subsubsection{The case of RDFS KGs}


\nd Now, rather than using an ontology reasoner, we will submit SPARQL queries to an SPARQL endpoint. Sometimes, SPARQL endpoints and/or RDF triple stores support some type of inference: the main purpose of the preprocessing is not to compute an initial classification but to activate the inference, if possible.\footnote{For example, in Virtuoso\footnote{\url{http://virtuoso.openlinksw.com}}, the `Reasoning mode' is enabled using the option \textsf{define input:inference "http://www.w3.org/2002/07/owl\#"}}

Unlike the OWL case, the instance retrieval problem for the class $C$ (Line~\ref{code:task1}) and the $d$-value retrieval problem (Line~\ref{code:task2}) can be achieved by means of a unique SPARQL query

\begin{lstlisting}[
    frame=single,
    numbers=left,
    basicstyle=\scriptsize\ttfamily,
    breaklines=true
]
PREFIX rdf: <http://www.w3.org/1999/02/22-rdf-syntax-ns#> .
PREFIX rdfs: <http://www.w3.org/2000/01/rdf-schema#> .
SELECT ?i ?v
WHERE {
   ?i rdf:type/rdfs:subClassOf* C .
   ?i d ?v 
}
\end{lstlisting}

%
\nd This query retrieves direct instances and assumes that the SPARQL point provides the direct instances of $C$ or one of its subclasses.

\subsection{Evaluating Type II sentences over classical knowledge bases}

\nd Algorithm~\ref{alg:type2ont} shows how to adapt the previous Algorithm~\ref{alg:1} to solve Type II sentences. Now, the algorithm requires more input parameters, as it needs an additional data property and an additional fuzzy datatype to characterize the fuzzy concept $F$. The main difference of the code is that now it is not only necessary to compute the membership degree of every individual $i$ to the fuzzy concept $G$ (Lines~\ref{code:GIni}--\ref{code:GFin}) but also to the fuzzy concept $F$ (Lines~\ref{code:FIni}--\ref{code:FFin}). Obviously, in the last part of the algorithm (Line~\ref{code:method2}), it invokes a method to solve Type II fuzzy quantified sentences rather than Type I sentences.



\begin{algorithm}[htbp]
\caption{Evaluating Type II quantified queries over knowledge bases}
\label{alg:type2ont}
\algorithmicrequire \ A knowledge base $\S$,  a classical concept $C$, functional numerical data properties $d_1, d_2$, fuzzy datatypes $L_1, L_2$ and and a fuzzy quantifier $Q$ 

\algorithmicensure \ $\alpha \in [0, 1]$ as the result of evaluating `$Q$ of 
$C \sqcap \exists d_1.L_1$ are $\exists d_2.L_2$'

%
\begin{algorithmic}[1]


\STATE $\S \gets \texttt{preprocess}(\S)$ 


\STATE $X \gets \texttt{getInstances}(\S, C)$ 

\STATE $F \gets \emptyset$ 

\STATE $G \gets \emptyset$ 

\FORALL{$i \in X$}   

        \STATE $v_{1i} \gets \texttt{getPropertyValue}(\S, i, d_1)$  \label{code:FIni}
        \IF {$v_{1i} \neq$\textsc{null} \AND } 
            \STATE $\alpha_{1i} \gets L_1(v_{1i})$
        \ELSE
            \STATE $\alpha_{1i} \gets 0$
        \ENDIF
        \STATE $F\gets F \cup \{\alpha_{1i} / i\}$ \label{code:FFin}
    
        \STATE $v_{2i} \gets \texttt{getPropertyValue}(\S, i, d_2)$ \label{code:GIni}
        \IF {$v_{2i} \neq$\textsc{null}} 
            \STATE $\alpha_{2i} \gets L_2(v_{2i})$
        \ELSE
            \STATE $\alpha_{2i} \gets 0$
        \ENDIF
        \STATE $G \gets G \cup \{\alpha_{2i} / i\}$ \label{code:GFin}

\ENDFOR

\RETURN $\texttt{solveTypeII}(Q, F, G)$ \label{code:method2}

\end{algorithmic}
\end{algorithm}

\subsection{Evaluating Type II sentences over fuzzy semantic knowledge bases}


\nd Algorithm~\ref{alg:4} shows how to evaluate Type II sentences over fuzzy ontologies. The main difference is that now we need a fuzzy ontology reasoner (such as fuzzyDL) to compute the instances of fuzzy concepts (Lines~\ref{code:task1f}--\ref{code:task2f}). Furthermore, the reasoner not only retrieves the instances but also their membership degrees. That is, given a fuzzy concept $Z$, it retrieves the set of pairs $\tuple{i, \alpha}$ such that $i$ belongs to $Z$ with degree greater or equal than $\alpha$, i.e.,
\[
X = \{ \tuple{\alpha/i} \mid \mathcal{O} \models Z(i) \geq \alpha \} \ .
\]
\nd In Line~\ref{code:task1f}, we have $Z = F_1$, 
whereas in  Line~\ref{code:task2f}, $Z = F_2$. 

As additional observations, note that although fuzzy ontology reasoners do not usually start by classifying the ontology, they typically implement some preprocessing, as fuzzyDL does~\cite{Bobillo16a,FuzzIEEE2020b}. Furthermore, as in the previous cases, the algorithm is agnostic to the method to solve a quantified sentence (Line~\ref{code:methodg}).

\begin{algorithm}
\caption{Evaluating Type II quantified queries over fuzzy ontologies}
\label{alg:4}
\algorithmicrequire \ A fuzzy ontology $\O$, a concept $C$, two functional numerical data properties $d_1, d_2$, two fuzzy datatypes $L_1, L_2$ and a fuzzy quantifier $Q$

\algorithmicensure \ $\alpha \in [0, 1]$ as the result of evaluating `$Q$ of $C \sqcap \exists d_1.L_1$ are $\exists d_2.L_2$'

%
\begin{algorithmic}[1]

\STATE $\O \gets \texttt{preprocess}(\O)$ 

\STATE $F_1 \gets C \sqcap \exists d_1.L_1$

\STATE $F_2 \gets \exists d_2.L_2$

\STATE $F  \gets \texttt{getInstances}(\S, F_1)$  \label{code:f1} 

\STATE $G  \gets \texttt{getInstances}(\S, F_2)$ \label{code:f2}  

\RETURN $\texttt{solveTypeII}(Q, F, G)$  \label{code:methodg}

\end{algorithmic}
\end{algorithm}

\begin{algorithm}
\caption{Evaluating Type II quantified queries over fuzzy ontologies}
\label{alg:5}
\algorithmicrequire A fuzzy ontology $\O$, two fuzzy concepts $F_1$ and $F_2$ and a fuzzy quantifier $Q$

\algorithmicensure $\alpha \in [0, 1]$ as the result of evaluating  `$Q$ of $F_1$ are $F_2$'

\begin{algorithmic}[1]

\STATE $\O \gets \texttt{preprocess}(\O)$ 

\STATE $F \gets \texttt{getInstances}(\O, F_1)$ \label{code:task1f} 

\STATE $G \gets \texttt{getInstances}(\O, F_2)$ \label{code:task2f} 

\RETURN $\texttt{solveTypeII}(Q, F, G)$ \label{code:methodf}

\end{algorithmic}
\end{algorithm}

Another important remark is that the algorithm can be adapted to receive directly two fuzzy concepts as an input, as shown in Algorithm~\ref{alg:5}. Again, 
Lines~\ref{code:task1f}--\ref{code:task2f} retrieve graded instances of $F_1$ and $F_2$, respectively.

Finally, note that Algorithm~\ref{alg:5} can also be applied to fuzzy KGs, provided that there is a method to retrieve graded instances of a given class. For example, given a class $C$ and a local KB, Fuzzy RDF system~\cite{Straccia09f} makes it possible to obtain a fuzzy set $X = \{ \langle \alpha / i \rangle \}$ as an answer to the query
\texttt{RETRIEVE ?i ?}$\alpha$\texttt{ WHERE ((?i rdfs:subclassof C) [}$\alpha$\texttt{])}.



\section{Implementation and validation}
\label{sec:implementation}


\paragraph{Implementation} 

To validate our approach, we have implemented \Name~(Quantified Query Semantic Solver).\footnote{\url{https://github.com/umberto-straccia/FuzzyQueries}} This tool enables the querying of both OWL~2 ontologies and KGs (via SPARQL endpoints) using various methods to evaluate fuzzy quantified sentences. Additionally, the prototype features a graphical user interface that allows users to easily select parameters and query terms.

\Name~has been implemented in Java, using \emph{JavaFX} for the graphical interface, and has the following dependencies:

\begin{itemize}

\item \emph{OWLAPI}~\cite{OWLAPI}\footnote{\url{http://owlcs.github.io/owlapi}} is a Java library to manage OWL~2 ontologies. In particular, we use it to load and access ontologies, but also as an interface to a reasoner;

\item \emph{Apache Jena}~\cite{Jena}\footnote{\url{http://jena.apache.org}} is a Java framework used to submit SPARQL queries to a remote SPARQL endpoint and manage RDF triples;

\item \emph{HermiT}~\cite{HermiT}\footnote{\url{http://www.hermit-reasoner.com}} is a reasoner for OWL~2 ontologies. It is mainly used to retrieve the instances of a class and the values of datatype properties of an individual;

\item \emph{Fuzzy OWL~2}~\cite{IJAR2011}\footnote{\url{www.umbertostraccia.it/cs/software/FuzzyOWL/}} is a Java library to manage fuzzy ontologies. In particular, \Name~uses it to parse the OWL~2 annotations describing fuzzy datatypes;

\item \emph{Datil}~\cite{Datil}\footnote{\url{http://webdiis.unizar.es/~ihvdis/Datil}} is a Java tool for fuzzy ontology learning. In particular, we use it to learn new fuzzy datatypes from the values of numerical data properties.

\end{itemize}


\nd \Name~employs a modular architecture to facilitate future extensions. For instance, we designed a unified data access interface that is currently specialized for OWL ontologies and SPARQL endpoints, but can be readily adapted for other data models. Consequently, integrating new evaluation methods for fuzzy quantified sentences, or swapping in different OWL~2 reasoners and SPARQL APIs, is a relatively straightforward process.

\begin{figure}[htbp]
    \centering
    \includegraphics[width=0.67 \linewidth]{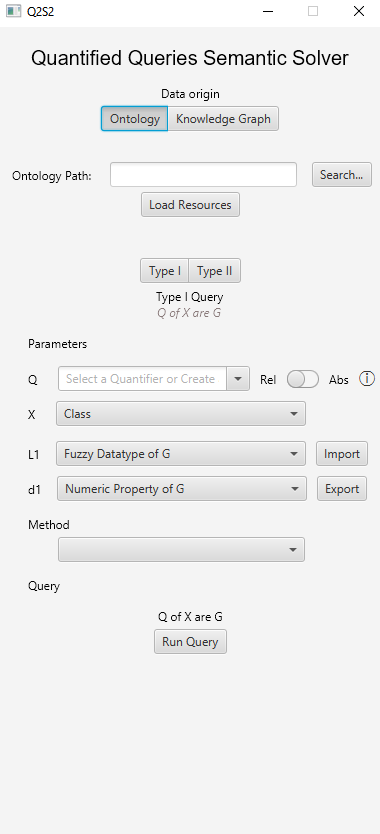}
    \caption{Graphical User Interface: Querying ontologies.}
    \label{fig:gui1}
\end{figure}

\begin{figure}[htbp]
    \centering
    \includegraphics[width=0.67 \linewidth]{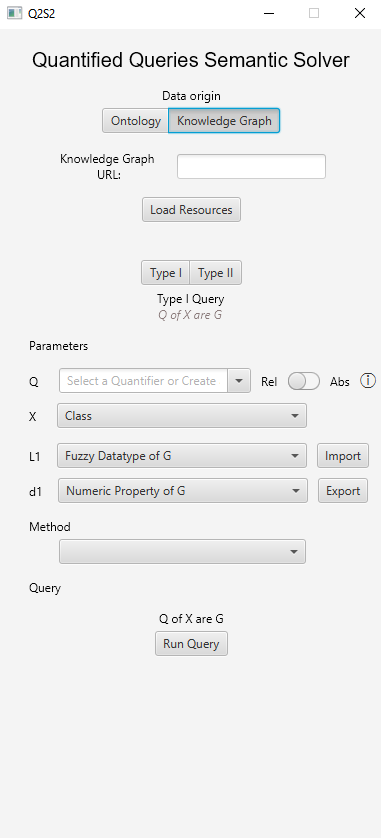}
    \caption{Graphical User Interface: Querying KGs.}
    \label{fig:gui2}
\end{figure}

\begin{figure}[htbp]
    \centering
    \includegraphics[width=\linewidth]{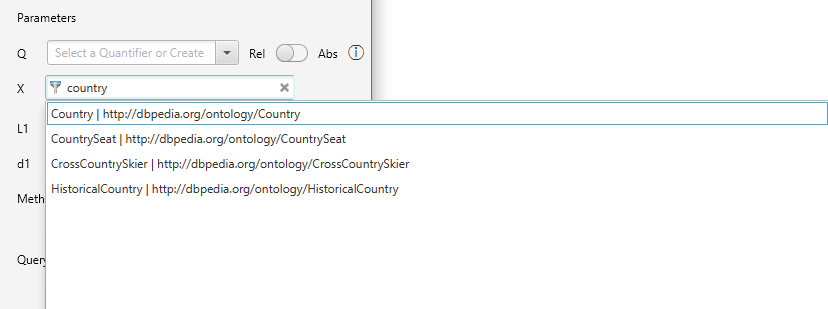}
    \caption{Graphical User Interface: autocomplete function.}
    \label{fig:gui3}
\end{figure}

From the graphical interface, the user can select an ontology or a SPARQL endpoint. Then, the user can select the type of query (I or II) and the query parameters. In particular, s/he can select classes, numeric properties, fuzzy datatypes, and fuzzy quantifiers (absolute or relative) from the definitions of the knowledge base. 
Furthermore, it is possible to select the methods to evaluate the query from a predefined list (at the moment of writing, it supports the methods in Section~\ref{sec:22}).
Figures~\ref{fig:gui1} and~\ref{fig:gui2} show the interfaces to query ontologies and KGs, respectively. The tool implements an auto--complete function to assist the user when typing the names of the entities, as shown in Figure~\ref{fig:gui3}.


Given that Fuzzy OWL~2 datatypes are not yet widely adopted in classical ontologies and KGs, \Name~ includes a feature to export property data values, so that the fuzzy datatypes can be learned from them using e.g.~Datil. Then, a local file with the Fuzzy OWL~2 datatypes can be seamlessly imported back into \Name.

\paragraph{Validation on an OWL ontology}

To validate \Name, we considered two use cases. It is well known that the different fuzzy quantification methods verify different logical properties. Rather than analyzing which is more appropriate, we cared about ensuring that the tool provides the expected result.

The first use case involves an OWL~ontology, namely Premierleague, which belongs to the SML-bench dataset,\footnote{\url{https://github.com/SmartDataAnalytics/SML-Bench}} a benchmarking framework for structured machine learning~\cite{SML-Bench}.

It is an ontology on the football domain with $10$ classes, $14$ object properties, $202$ data properties, $11,859$ individuals, and $2,130,363$ logical axioms. The expressivity of the ontology is the DL
$\mathcal{ALEH}(\mathbf{D})$. 

Essentially, it makes it possible to represent sets of statistics (members of the \textsf{Action} class) of Premier League matches (\textsf{Match} class). For example, \textsf{action1008910930} is an \textsf{Action}, corresponding to a player (\textsf{Petrov Martin}) of a team (\textsf{Bolton Wanderers}) on a certain match (\textsf{match109}).\footnote{It corresponds to Bolton Wanderers: Chelsea (2011-10-02).} Examples of the relationships of \textsf{action1008910930} are a \textsf{passes\_efficiency} of $0.68$ or a 
\textsf{shot\_efficiency} of $0$.

We used Datil to fuzzify some data properties, using k-means algorithm to create $5$ clusters. 
For each data property, we considered its property values and fuzzify them into $5$ fuzzy datatypes. In particular, we fuzzified \textsf{shot\_efficiency}.

Then, we evaluated several queries. For the sake of concrete illustration, we will consider here an example of each case. The evaluation was performed on a laptop computer running Windows 7 64-bits, Intel Core i7-8550U 1.8 GHz, 16~GB RAM.
Table~\ref{tab:results} shows the preprocessing time, the times needed to evaluate the first and the second query, and the result of the evaluation. Numbers were rounded to $3$ decimals.

As an example of Type I query, we used \Name~to check whether in few of the matches, the shot\_efficiency is high. To do so, we evaluated the sentence 
\begin{quote}
    Few Actions have a shot\_efficiency which is HighShot\_efficiency (Q1) \ ,
\end{quote}
\nd where \textsf{Action} is a crisp concept, \textsf{shot\_efficiency} a data property, and \textsf{HighShot\_efficiency} a fuzzy datatype, and the quantifier ``few'' is defined as $\mathtt{left}(0.2, 0.5)$.

\begin{figure}[htbp]
     \centering
     \includegraphics[width=0.67 \linewidth]{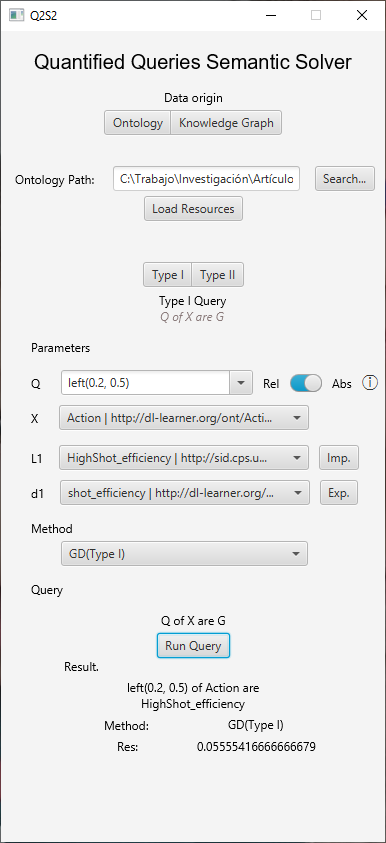}
     \caption{Graphical User Interface: Type I query over Premier League ontology.}
     \label{fig:gui4}
 \end{figure}

For example, using the GD method, the result of the query is $0.056$ (rounded to $3$ decimals), as shown in Figure~\ref{fig:gui4}. The preprocessing time and the time needed to solve the first time were relatively high, around $20$ and $6$ seconds, respectively. However, we must recall that, because of the high number of individuals and axioms, it is a challenging ontology and they are computed just once.
Interestingly, for subsequent queries, the evaluation time was about $0.1$ seconds.
%
%
Another query we considered is
\begin{quote}
    Most Actions with HighShot\_efficiency have a HighPass\_efficiency (Q2),
\end{quote}
\nd where `most' is defined as $\mathtt{right}(0.5, 1)$.

\paragraph{Validation on an RDFS knowledge graph}

We also evaluated \Name~ by querying the very large DBpedia KG. In particular, we focused on a geographical domain, querying about the \textsf{population}, \textsf{populationDensity}, \textsf{elevation}, \textsf{area}, etc. of cities (\textsf{City} class). DBpedia includes $27,567$ cities, but some of them have unknown values for some of the attributes. For example, it includes the population density of $10,204$ cities and the elevation of $16,980$ cities.

Again, we fuzzified the data properties, computing $5$ fuzzy datatypes for each of them, using Datil. It is worth to note that we discarded some outliers to have more reliable definitions. For example, for the \textsf{populationDensity} property we considered an outlier (Buenos Aires' population, $3,121,707$) and for the \textsf{elevation} property we considered another one (Gunnaur's elevation, $118,88$).

\begin{figure}[htbp]
     \centering
     \includegraphics[width=0.67 \linewidth]{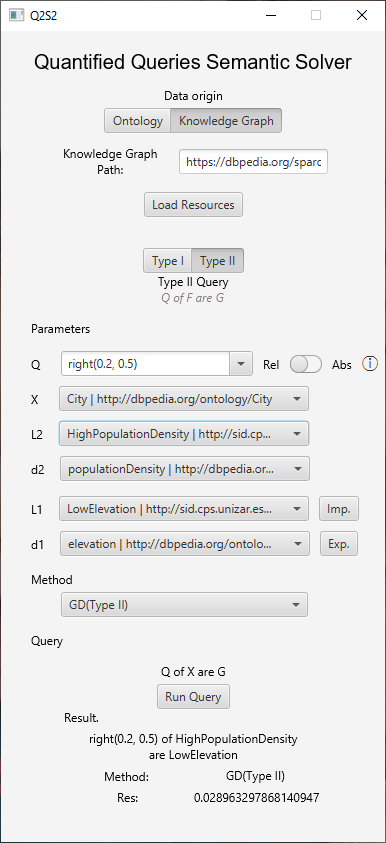}
     \caption{Graphical User Interface: Type II query over DBpedia knowledge graph.}
     \label{fig:gui5}
 \end{figure}

We considered several Type-I and Type-II queries. For the sake of concrete illustration, we consider the queries
\begin{quote}
    Few Cities are HighPopulationDensity (Q3)
\end{quote}
\begin{quote}
    At least some Cities with HighPopulationDensity are LowElevation (Q4),  
\end{quote}
\nd where `at least some' is defined as $\mathtt{right}(0.2, 0.5)$. For example, using the GD method to solve Q4, the result was $0.029$, as shown in Figure~\ref{fig:gui5}. In this case, as shown in Table~\ref{tab:results}, the preprocessing and the evaluation time are much smaller, and the first query over a KG is not significantly different from the subsequent ones. There are also no significant differences between the methods.

\begin{table*}[thbp]
\begin{center}
\scalebox{0.7}{
\begin{tabular}{|ccc|c|ccc|ccc|}%
\hline
 & & & Preprocess & \multicolumn{3}{|c|}{GD} & \multicolumn{3}{|c|}{Zadeh} \\ 
Dataset	&	Query	&	Type	&	Time	&	Time (1st)	&	Time (2nd)	&	Result 	&	Time (1st)	&	Time (2nd)	&	Result \\
\hline
Premierleague	&	Q1	&	Type I	&	20160	&	5515	&	145	&	0.056 &	5267	&	86	&	0.0  \\
Premierleague	&	Q2	&	Type II	&	20160	&	6877	&	124	&	0.193	&	5561	&	144	&	0.217	\\
DBpedia	&	Q3 &	Type I	&	47	&	1380	&	798	&	1.0	&	996	&	1095	&	1.0	\\
DBpedia	&	Q4	&	Type II	&	47	&	1407	&	1417	&	0.029	&	1638	&	1353	&	0.052	\\
\hline
\end{tabular}
}
\caption{Evaluation of queries Q1--Q4 using GD and Zadeh's methods. Times are shown in ms. }
\label{tab:results}
\end{center}
\end{table*}

\color{black}


\section{Related work}
\label{sec:relatedWork}

\paragraph{Fuzzy quantification and ontologies}

D. Sánchez and A. Tettamanzi~\cite{Sanchez05a,Sanchez06,Sanchez05b} were the first authors to consider fuzzy quantifiers (beyond $\exists$ and $\forall$) and fuzzy ontologies. More precisely, they proposed a fuzzy extension of the Description Logic $\mathcal{ALCQ}\mathbf{D)}$, with fuzzy concepts of the form $Q R.C$, denoting the fuzzy set of elements such that $Q$ of the successors via the fuzzy property $R$ are members of the fuzzy concept $C$, which can be seen as Type I or Type II sentences. The authors also discuss how to solve the concept satisfiability problem, which includes using the GD method to evaluate the fuzzy quantified sentences. Instead, our work is focused on the computation of the degree to which a fuzzy quantified sentence holds. Furthermore, our approach can be applied to classical ontologies and KGs and discusses the representation using Fuzzy OWL~2.

F. Bobillo and U. Straccia supported some fuzzy quantifiers in fuzzy ontologies~\cite{BobilloASOC2013}. In particular, they were interested in representing quantifier-based OWA aggregation concepts, where the weights are directly computed from a regularly increasing quantifier. Such concepts can be represented in the language Fuzzy OWL~2~\cite{IJAR2011} and are supported by the reasoner fuzzyDL~\cite{KBS2016}. However, this approach does not evaluate Type I or Type II sentences.

F. A. Lisi and C. Mencar proposed a granular computing approach to enrich classical OWL~2 ontologies~\cite{Lisi17,LisiFI18}. 
Their method uses fuzzy clustering and SPARQL to process numerical properties within OWL ontologies, transforming precise data into fuzzy sets and integrating them as new, interpretable assertions within the ontology. The authors also discuss how to represent quantified sentences such as `$Q$ of $C$ have a value of property $d$ which is $L$'. To illustrate the method, a system for building a granular view of individuals over an OWL ontology called GranulO was implemented~\cite{Lisi16}.


\paragraph{Fuzzy quantification and KGs}

Regarding knowledge graphs, O. Pivert et al.~\cite{PivertFUZZIEEE2017} discussed how to solve Type I and Type II sentences over fuzzy RDF graphs with  statements of the form $\langle s, p, o, \alpha \rangle$. To represent the quantified queries, they proposed an extension of the FURQL (Fuzzy RDF Query Language) language, a fuzzy extension of SPARQL~\cite{PivertFUZZIEEE2016}. The approach is implemented in a tool called SURF (SPARQL with fUzzy quantifieRs for rdF data), built on top of Jena Semantic Web Java Framework, although the authors needed to implement a new algorithm to compute the satisfaction degree of individual to a FURQL query. The implementation is restricted to three methods (Zadeh's method and two Yager's methods) involving relative quantifiers. The main differences with our work are that we support standard RDF triples and standard SPARQL queries to retrieve the individuals from the graph. Furthermore, we support fuzzy datatypes and the more complex GD method to evaluate the sentences.

G. Li et al.~\cite{LiIJIS19} studied fuzzy quantified graph pattern matching given fuzzy RDF statements of the form $\langle s / \alpha_1, p / \alpha_2, o / \alpha_3 \rangle$. The authors implemented Zadeh's method to solve Type I and Type II sentences. An example of the supported queries, provided by the authors, is ``Most of the recent films that actor $x$ starred in, are directed by young directors'', where both ``recent'' and ``young'' are fuzzy sets. Instead, we do not assume that triples are fuzzy and we support more methods to solve quantified queries. 
Moreover, their method lacks the use of fuzzy datatypes for representing both the fuzzy linguistic terms in the query and the quantifiers.

%

\section{Conclusions and future work}
\label{sec:conclusions}

\nd This paper introduces a highly adaptable framework for evaluating Type I and Type II fuzzy quantified queries across both standard and fuzzy ontologies. A major strength of the proposal is its agnostic design, allowing it to operate independently of the quantifier type, evaluation method, or data source (e.g., OWL ontologies or RDFS KGs). For classical data sources (semantic knowledge bases), the framework integrates fuzzy datatypes so that standard semantic reasoners can still be employed. For fuzzy data sources, it leverages specialized fuzzy tools. To facilitate ongoing research, a publicly accessible implementation of the framework called \Name~ has been released. We confirmed on two uses cases that the evaluation time is appropriate 
even on large ontologies and KGs.

In future research, we aim to expand the proposed framework to support the evaluation of arbitrary nested quantified sentences. This enhancement will enable the system to process complex, multi-layered queries, such as `\emph{Few} expensive hotels feature \emph{many} low-price amenities'. 



\section*{Acknowledgment}

\nd I. Huitzil, U. Straccia, and F. Bobillo were partially supported by the I+D+i project PID2024-159530OB-I00, funded by MCIN/AEI/10.13039/501100011033. 
I.\ Huitzil and F. Bobillo were also supported by the project UZ2024-IyA-02, funded by University of Zaragoza.




\bibliographystyle{abbrv}

\end{document}